\documentclass[10pt,twocolumn,letterpaper]{article}

\usepackage{iccv}
\usepackage{times}
\usepackage{epsfig}
\usepackage{graphicx}
\usepackage{amsmath}
\usepackage{amssymb}
\usepackage{booktabs}
\usepackage{subfigure}
\usepackage{multicol}
\usepackage{multirow}
\usepackage{lipsum}

\usepackage[accsupp]{axessibility} 

\def\bb{\mathbf{b}}

\def\x{\mathbf{x}}

\def\y{\mathbf{y}}

\def\vv{\mathbf{v}}

\DeclareMathOperator{\B}{B}

\DeclareMathOperator{\Adj}{A}

\DeclareMathOperator{\F}{F}
\DeclareMathOperator{\Hh}{H}

\DeclareMathOperator{\Pj}{P}

\DeclareMathOperator{\Vm}{V}

\DeclareMathOperator{\Y}{Y}

\DeclareMathOperator{\Um}{U}
\DeclareMathOperator{\Z}{Z}
\DeclareMathOperator{\W}{W}

\newcommand{\R}{\ensuremath{\mathbb{R}}}

\usepackage[pagebackref=true,breaklinks=true,letterpaper=true,colorlinks,bookmarks=false]{hyperref}
\iccvfinalcopy 

\def\iccvPaperID{****} 

\ificcvfinal\fi

\begin{document}

\title{CaT: Weakly Supervised Object Detection with Category Transfer}

\author{Tianyue Cao$^1$\quad 
Lianyu Du$^1$\quad 
Xiaoyun Zhang$^1$\thanks{Xiaoyun Zhang is the corresponding author.} \quad 
Siheng Chen$^{1,2}$\quad 
Ya Zhang$^{1,2}$\quad 
Yan-Feng Wang$^{1,2}$\\
Cooperative Medianet Innovation Center, Shanghai Jiao Tong University$^1$\quad Shanghai AI Laboratory$^2$\\
{\tt\small \{vanessa\_, dulianyu, xiaoyun.zhang, sihengc, ya\_zhang, wangyanfeng\}@sjtu.edu.cn}
}

\maketitle

\newcommand\blfootnote[1]{%
\begingroup
\renewcommand\thefootnote{}\footnote{#1}%
\addtocounter{footnote}{-1}%
\endgroup
}

\ificcvfinal\thispagestyle{empty}\fi

\begin{abstract}
    A large gap exists between fully-supervised object detection and weakly-supervised object detection. To narrow this gap, some methods consider knowledge transfer from additional fully-supervised dataset. But these methods do not fully exploit discriminative category information in the fully-supervised dataset, thus causing low mAP. To solve this issue, we propose a novel category transfer framework for weakly supervised object detection. The intuition is to fully leverage both visually-discriminative and semantically-correlated category information in the fully-supervised dataset to enhance the object-classification ability of a weakly-supervised detector. To handle overlapping category transfer, we propose a double-supervision mean teacher to gather common category information and bridge the domain gap between two datasets. To handle non-overlapping category transfer, we propose a semantic graph convolutional network to promote the aggregation of semantic features between correlated categories. Experiments are conducted with Pascal VOC 2007 as the target weakly-supervised dataset and COCO as the source fully-supervised dataset. Our category transfer framework achieves 63.5\% mAP and 80.3\% CorLoc with 5 overlapping categories between two datasets, which outperforms the state-of-the-art methods. Codes are avaliable at \url{https://github.com/MediaBrain-SJTU/CaT}\blfootnote{This work was supported in part by  Chinese National Key R\&D Program (2019YFB1804304), National Natural Science Foundation of China (61771306),State Key Laboratory of UHD Video and Audio Production and Presentation, Shanghai Key Laboratory of Digital Media Processing and Transmissions(STCSM 18DZ2270700) and 111 plan(BP0719010).}.
\end{abstract}

\vspace{-5mm}
\section{Introduction}

\begin{figure}[t]
\begin{center}
   \includegraphics[width=\linewidth]{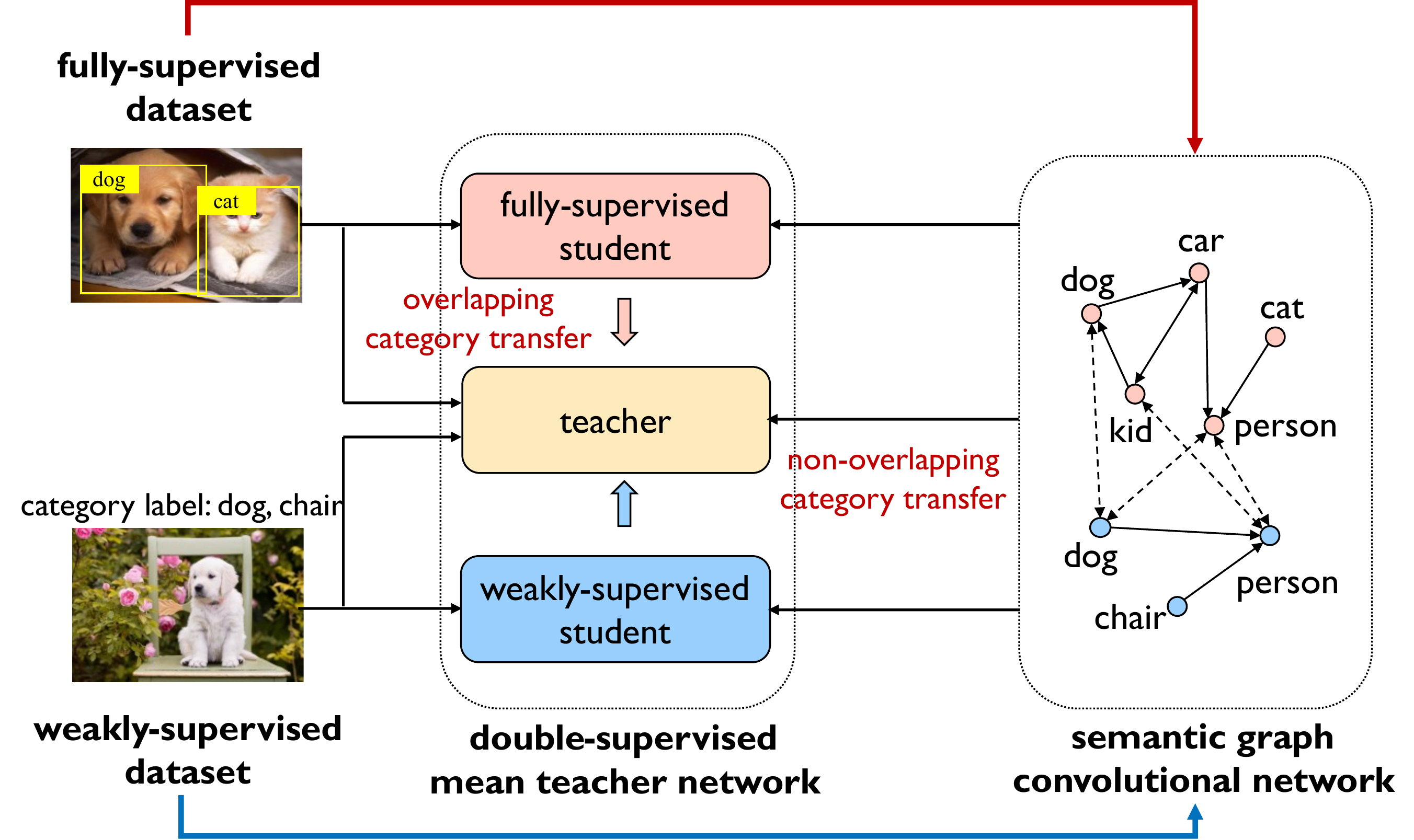}
\end{center}
   \caption{The sketch map of our category transfer framework. We use the double-supervised mean teacher network to transfer overlapping category and objectness knowledge; and use the semantic graph convolutional network to transfer non-overlapping category knowledge. The final prediction is generated by the teacher.}
   \vspace{-5mm}
\label{fig:frame}
\end{figure}

Object detection is one of the most fundamental tasks in computer vision~\cite{odsurvey}. In the past decade, based on deep neural networks, many methods~\cite{cornernet,yolov3,fasterrcnn} have achieved great success. However, most methods follow the fully supervised setting, which requires a huge number of high-quality annotations, including the precise bounding boxes of objects and their corresponding category labels. This setting usually costs extensive amount of time and resources to acquire such annotations. To reduce the annotation cost, weakly supervised object detection (WSOD) are proposed~\cite{wsddn,pcl,cmil} to train detection models with only image-level category labels. However, the lack of bounding-box-level supervision leads to significant issues, such as instance ambiguity and low-quality proposals. Therefore, a large performance gap still exists between fully-supervised object detection (89.1\% mAP, SOTA~\cite{cascade}) and weakly-supervised object detection (56.8\% mAP, SOTA~\cite{casd}).

To narrow this gap, some previous methods consider knowledge transfer from additional data. There are two main approaches: the objectness transfer approach and the semi-supervised approach. For example,~\cite{25,20,eccv2020} train a generic object detector on source data and apply it to target data; however, this objectness transfer methods ignores the category information in the source dataset, causing the decline of classification. ~\cite{lsda,24,17,22} follow the semi-supervised setting with partial fully annotated data, and transform an image classifier to an object detector. Such a semi-supervised method leverages both box and category information, but it usually can not solve the domain gap between datasets, especially the category inconsistency between the source and target datasets. Moreover, the correlations between categories have not been exploited. Therefore, the lack of exploiting category information still limit the empirical performance of the target dataset.

To solve those issues, this work specifically considers category transfer; that is, leverage  both visually-discriminative and semantically-correlated category information in the fully-supervised dataset to enhance the discriminative ability of a weakly-supervised detector. Based on whether the categories are shared in both the fully-supervised and the weakly-supervised datasets, the category transfer includes overlapping category transfer,  where  the fully-supervised and the weakly-supervised datasets share the same categories, and non-overlapping category transfer, where two datasets have different, yet correlated categories.

To realize overlapping category transfer, we propose a double-supervision mean teacher network. The double-supervision mean teacher network adopts the similar structure of mean teacher method~\cite{meanteacher}, but works with two students, each of which is supervised by either fully-supervised or weakly-supervised datasets. Our teacher thus can gather overlapping category information from both fully-supervised and weakly-supervised students, leading to better  discriminative ability. Moreover, the mean teacher structure could bridge the domain gap between two datasets, including differences in image and category distributions.

\begin{figure}[t]
\begin{center}
   \includegraphics[width=0.8\linewidth]{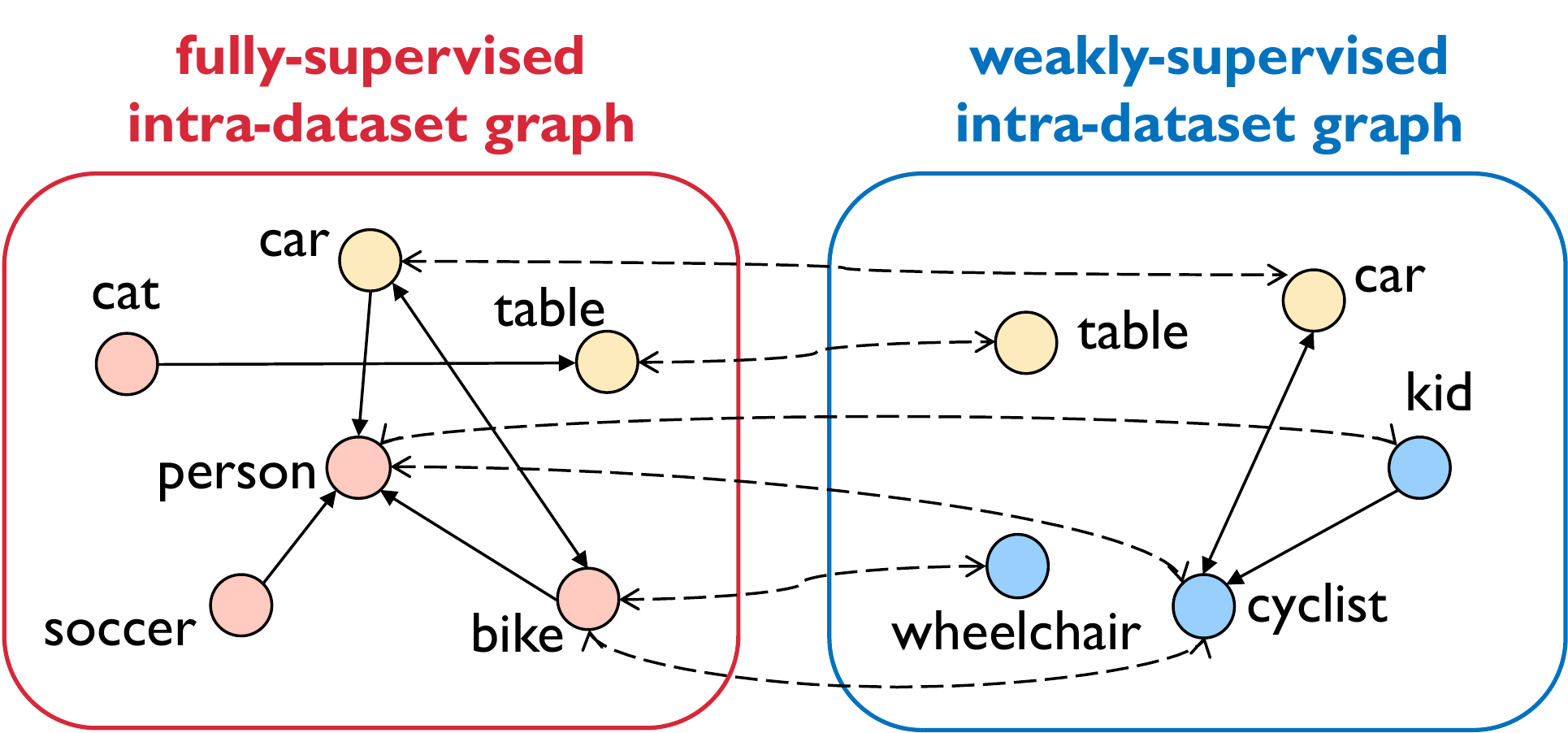}
\end{center}
   \caption{An example of category label graph. Red, blue and yellow nodes represent fully-supervised, weakly-supervised and common categories separately. The inter-dataset graph edges are built according to common categories (\emph{``car(fully)"} $\leftrightarrow$ \emph{``car(weakly)"}) and category relationships. The relationships include subclass relationship (\emph{``person(fully)"} $\leftrightarrow$ \emph{``kid(weakly)"}), including relationship (\emph{``cyclist(weakly)"} $\leftrightarrow$ \emph{``person(fully)"}), and similarity relationship (\emph{``bike(fully)"} $\leftrightarrow$ \emph{``wheelchair(weakly)"}). Each intra-dataset graph is a digraph, and the inter graph is a bigraph with bi-directional edges. Best viewed in color.}
\label{fig:categories}
   \vspace{-3mm}
\end{figure}

To achieve non-overlapping category transfer, we propose a semantic graph to model the correlations among all the categories in both fully-supervised and weakly-supervised datasets. The intuition is that even two categories are different, they may be highly-correlated and their corresponding category information can be strategically transferred according to the correlation strength. Figure~\ref{fig:categories} illustrates a toy example of a semantic graph. The category~\emph{kid} in the weakly-supervised dataset does not exactly match with the category~\emph{person} in the fully-supervised dataset, but they are semantically related. Based on this semantic graph, we use graph convolutional networks to exploit non-overlapping category information and provide semantic guidance for object classification.  

Overall, we propose an end-to-end framework; see Figure \ref{fig:frame}. It includes a Faster R-CNN~\cite{fasterrcnn} as the backbone network, a double-supervised mean teacher network for overlapping category transfer, and a semantic graph convolutional network for non-overlapping category transfer. This framework can fill the domain gap by using a mean teacher structure, and fully exploit category information by aggregating semantic features over a semantic graph. Compared with objectness transfer approaches, we transfer category information from a fully-supervised dataset to improve the classifier. Compared with approaches under the semi-supervised setting \cite{lsda,24,17,22}, our approach applies a double supervised mean teacher to solve domain gap between datasets, and uses a semantic graph convolutional network to fully exploit the correlations between categories, leading to semantic transfer of non-overlapping categories. Our method outperforms the state-of-the-art methods in WSOD, and achieves results competitive to FSOD baseline.

We summarize our main contributions as:

$\bullet$ We propose a novel category transfer framework for WSOD, which specifically handles the issues on category transfer, including the domain gap problem in overlapping category transfer and information aggregation issue in non-overlapping category transfer.
	
$\bullet$ We propose a novel double-supervised mean teacher network to handle overlapping category transfer. This network gathers common category information and bridges the domain gap between two datasets.
	
$\bullet$ We propose a novel semantic graph convolutional network to tackle non-overlapping category transfer. This network promotes the aggregation of semantic features between correlated categories.

$\bullet$ We conduct extensive experiments and show that the proposed method outperforms the state-of-the-art weakly-supervised object detection methods and is competitive to fully-supervised object detection baseline on benchmarks.

\section{Related Works}

\textbf{Weakly-Supervised Object Detection.} Recent works usually formulate WSOD as a MIL problem. WSDDN\cite{wsddn} is the first end-to-end MIL deep neural network in WSOD. Inspired by WSDDN, many methods are proposed. \cite{oicr} integrates WSDDN and a multi-stage online instance classifier refinement (OICR) algorithm into a simple deep network. \cite{pcl} use an iterative process, Proposal cluster learning (PCL), to learn refined instance classifiers based on OICR. \cite{cmil} introduces continuation multiple instance learning (C-MIL) by smoothing the original loss function to solve the part domination problem. \cite{wsod2} uses top-down confidence and bottom-up evidence with an adaptive training mechanism to distill box boundary knowledge. Follow-up works further improve the performance by self-training \cite{wetectron}, self-distillation \cite{wsod2} \cite{casd}, etc.

\textbf{Weakly-Supervised Object Detection with Knowledge Transfer.} Transfer learning is typically used in domain adaptation (DA) \cite{da1}. It can bridge different datasets\cite{meanteacher}, categories\cite{opda}, or even tasks\cite{lsda,task2}, which reduces the cost of training and leverages the knowledge from another dataset/domain/task. In this paper, we focus on knowledge transfer in weakly-supervised object detection. Additional data are usually used to provide auxiliary prior information for detection. \cite{17,18} use the word embedding of category label text to represent the semantic relationship between classes, and the dependency of categories is used to assist object detection. \cite{19,20,21} transfer the model learned in source domain to target domain. Also, some works \cite{22,lsda,17} use weight prediction to effectively transform an image classifier into an object detector. Recent studies \cite{23,24,25,26} raise the idea of sharing the general knowledge learned in source domain. The knowledge can be object predictor \cite{23,26}, object candidate region \cite{24}, or general boundary box regression \cite{25,eccv2020}. However, categories in two datasets are usually not the same, most of the existing methods do not work under this situation. In this paper, we combine WSOD with mean teacher framework to fully leverage knowledge from both a public fully-supervised dataset and a weakly-supervised dataset, which can also solve the category mismatch problem.

\textbf{Knowledge-Guided Graph Reasoning.} Graph reasoning is proved to be effective in many tasks, including image classification \cite{gnncls,iccv2019}, object detection \cite{reasoningrcnn,universalrcnn}, human skeleton-based action recognition \cite{maosen2} or motion prediction \cite{maosen1}, etc. These methods model domain knowledge as a graph to transfer knowledge based on category dependency, object spatial relationship, or object semantic relationship. Some classification models \cite{gnncls,iccv2019} build a category dependency graph based on the dataset statistical information. \cite{reasoningrcnn} uses knowledge graph to discover most relative categories for feature evolving. Our method not only uses a semantic graph for both fully-supervised and weakly-supervised datasets to reason category dependencies but also transfers category similarity or dependency knowledge between two datasets.

\vspace{-1mm}
\section{Category Transfer Framework}

\textbf{Problem formulation.}
Mathematically, given a weakly-supervised dataset $\mathcal{D}_w$, each image $\x_w$ in the dataset has an image-level category label $\y_w$ with $\y_w \in \R^{C_w}$, where $C_w$ is the number of weakly-supervised dataset categories.  Similarly, for a fully-supervised dataset $\mathcal{D}_f$, each image $\x_f$ has the instance-level annotations, including  the bounding boxes $\B_{f} = \{\bb^{(i)}_f\}_{i=1}^{r}$ and their corresponding category labels $\Y_f = \{\y^{(i)}_f\}_{i=1}^{r}$, where $r$ is the number of instances in an image and $\y^{(i)}_f \in \mathbb{R}^{C_f}$ is the category label of the $i$th instance with $C_f$, the number of fully-supervised dataset categories.  We aim to train a model $\mathcal{M}$ on $\mathcal{D}_w$ with knowledge transfer from $\mathcal{D}_f$. For each testing image $\x$, the model can output instance-level detection, including  estimated bounding boxes and category estimations; that is $\widehat{\B}, \widehat{\Y} = \mathcal{M}(\x)$.

\begin{figure*}[t]
\begin{center}
   \includegraphics[width=\linewidth]{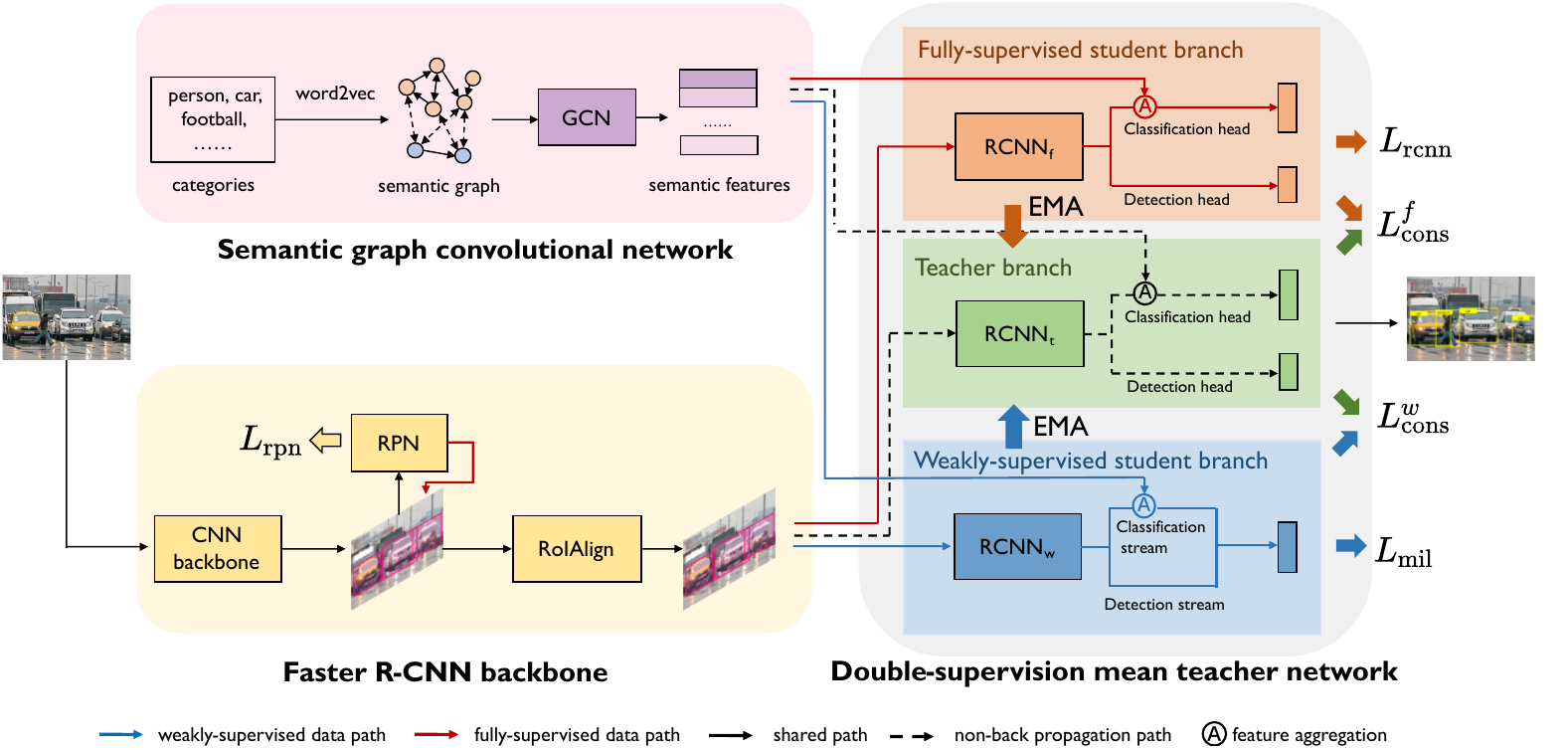}
\end{center}
   \caption{Overall architecture of our category transfer framework. Image region features are forward to the double-supervision mean teacher, which has three branches: the fully-supervised student branch, the weakly-supervised student branch, and the teacher branch. The semantic features are updated by the semantic graph convolutional network and fused to visual features for classfication. The training process in one iteration contains two forward steps (solid lines), one backpropagation step (dotted lines), and one exponential moving average (EMA) step.}
   \vspace{-5mm}
\label{fig:framework}
\end{figure*}

\textbf{Overview.} To design such a model, we specifically consider category transfer; that is, leverage the category information for object recognition in a fully-supervised dataset to enhance the discriminative ability of a detector, which is trained on a weakly-supervised dataset. Category transfer enables the discriminative ability transfer from a fully-supervised dataset to a weakly-supervised detector, leading to better detection performance.

As shown in Figure \ref{fig:framework}, the proposed category transfer framework includes three parts: the backbone network, the double-supervision mean teacher network (Section \ref{mt}) and the semantic graph convolutional network (Section \ref{sgcn}). For an input image, our backbone network follows the backbone of Faster R-CNN~\cite{fasterrcnn}, extracts visual features and generates the regions of proposals. The double-supervision mean teacher network takes the pooled visual features in each region and estimates the bounding boxes and classes by leveraging the bounding box and overlapping category information from both fully-supervised and weakly-supervised datasets. To further enable non-overlapping category transfer, the semantic graph convolutional network exploits the semantic correlations among categories and outputs the semantic features for each category. To classify each region, we aggregate the semantic features produced by the semantic graph convolutional network and the visual features from the R-CNN output of the double-supervision mean teacher network to generate final category features, boosting the recognition ability.

\subsection{Double-Supervision Mean Teacher Network} \label{mt}
The proposed double-supervision mean teacher network follows the traditional mean teacher, yet with two different student branches. This novel architecture can bridge the domain gap and estimate the bounding boxes and classes via overlapping category transfer.

\textbf{Network architecture.} This network consists of a fully-supervised student branch, a weakly-supervised student branch, and a teacher branch. Each of the three branches takes the pooled visual features for proposal regions produced by the backbone network as the input and updates the visual feature through a few convolutional layers; that is,  given the pooled visual feature $\F^{p}$, the visual features after convolutional layers in Branch $*$ is $\F_*^{o} = {\rm RCNN}_*(\F^{p}) \in \mathbb{R}^{r \times d}$,  where $r$ is the number of region proposals, $d$ is the dimension of features, RCNN$_*(\cdot)$ is the regional convolutional neural network and $*$ is the branch index with $f$ for the fully-supervised branch, $w$ for the weakly-supervised branch and $t$ for the teacher branch.

Taken fully-supervised region features $\F_f^{o}$ as input, the fully-supervised student branch uses a regression head and a classification head to  output the estimated bounding boxes $\widehat{\B}_f = \{\widehat{\bb}_f^{(i)}\}_{i=1}^{r}$ and the corresponding categories $\widehat{\Y}_f = \{\widehat{\y}_f^{(i)}\}_{i=1}^{r}$, respectively.  The outputs are supervised by the ground-truth bounding boxes and the the corresponding category labels in the fully-supervised dataset. Note that the estimated label is $\widehat{\y}_f^{(i)} \in \mathbb{R}^{C_f}$, where $C_f$ is the number of categories in the fully-supervised dataset. The architecture of this branch is the same as class-agnostic Faster R-CNN~\cite{fasterrcnn} after RoIAlign layer.

\begin{figure}[t]
\begin{center}
   \includegraphics[width=\linewidth]{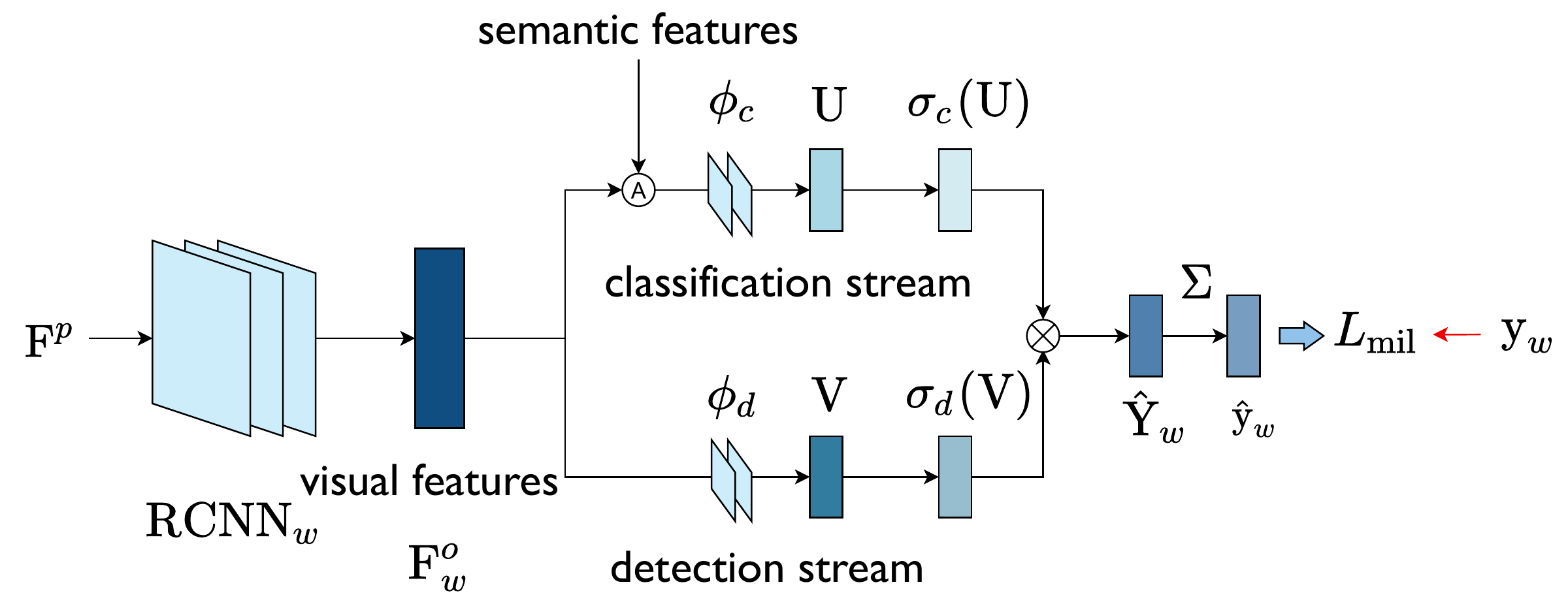}
\end{center}
   \caption{Architecture of the weakly-supervised student branch.}
\label{fig:wsbranch}
\vspace{-3mm}
\end{figure}

The weakly-supervised student branch trains a multiple-instance-learning (MIL)-based network inspired by~\cite{wsddn}. Taken weakly-supervised region features $\F_w^{o}$ as input, this branch outputs the image-level category prediction $\widehat{\y}_w \in \mathbb{R}^{C_w}$, which is supervised by the image-level category labels in the weakly-supervised dataset. Figure~\ref{fig:wsbranch} illustrates its architecture. The output feature of the RCNN is input to a classification stream and a detection stream. The classification stream uses a fully-connected layer $\phi_c$ to map $\F_w^{o} \in \mathbb{R}^{r \times d}$ to $\Um \in \mathbb{R}^{r \times C_w}$, where $r$ is the number of region proposals, $C_w$ is the number of categories in the weakly-supervised dataset. We then apply the softmax operation along the category (second) dimension of $\Um $ and obtain the category prediction scores $\sigma_c(\Um) \in \mathbb{R}^{r \times C_w}$ for $r$ proposals. Meanwhile, the detection stream uses another fully-connected layer $\phi_d$ to map $\F_w^{o}$ to $\Vm \in \mathbb{R}^{r \times C_w}$ and applies the softmax operation along the proposal (first) dimension of $\Vm$ to obtain the detection score $\sigma_d(\Vm) \in \mathbb{R}^{r \times C_w}$ for $r$ proposals. Finally, we apply the element-wise multiplication between the results of two streams to get the image-level category prediction.

The teacher branch accumulates knowledge from two student branches by exponential moving average (EMA)~\cite{meanteacher} without training. It takes both fully-supervised and weakly-supervised region features as input, and outputs the final box predictions $\widehat{\B}$ and category predictions $\widehat{\Y}$. The teacher branch has a similar architecture as the fully-supervised student branch and also includes the regression head and the classification head. The difference is that the number of output categories of the teacher branch follows the categories in the weakly-supervised dataset. To transfer the regression ability, we take the exponential moving average of the weights in the regression head of the fully-supervised student branch as the weights in the regression head of the teacher branch.  To transfer the classification ability, we take the exponential moving average of the weights in the classification stream of the weakly-supervised student branch as well as the weights of overlapping categories in the fully-supervised student branch to be the weights in the classification head of the teacher branch.

\textbf{Loss function.}
The overall loss function includes three parts: the weakly-supervised MIL loss $L_{\rm mil}$, which is an image-level cross-entropy classification loss~\cite{wsddn}; the fully-supervised loss $L_{\rm full}$, which is same as Faster R-CNN~\cite{fasterrcnn}, including RPN loss $L_{\rm rpn}$ and R-CNN loss $L_{\rm rcnn}$; and the consistency loss  to promote consistency between the teacher and two student branches; that is,
$L_{\rm  cons} =  L_{\rm cons}^{f} + L_{\rm cons}^{w},$
where the fully-supervised  and weakly-supervised consistency losses are
$
L_{\rm cons}^{f}  =  L_{\rm box} (\widehat{\bb}_{f}, \widehat{\bb}) +  L_{\rm cls} (\widetilde{\y}_{f}, \widetilde{\y}),
L_{\rm cons}^{w} = L_{\rm cls}(\widehat{\y}_{w}, \widehat{\y}),
$
respectively, with $\widehat{\bb}$ the estimated bounding box of the teacher branch, $\widetilde{\y}$ the classification score of the teacher branch for the overlapping categories, $\widehat{\y}$ the classification score of the teacher branch for all the categories,  $L_{\rm cls}$ is the smooth-$\ell_1$ loss for classification consistency and $L_{\rm box}$ is the mean square error for box regression consistency. 

The overall loss function is
$L = L_{\rm mil} + \lambda_{\rm full} L_{\rm full} + \lambda_{\rm cons} L_{\rm cons},
$
where $\lambda_{\rm full}$ and $\lambda_{\rm cons}$ are the hyperparameters to balance the weakly-supervised, fully-supervised, and consistency losses.

\subsection{Semantic Graph Convolutional Network} \label{sgcn}
To leverage non-overlapping categories in the fully-supervised dataset, we propose a novel semantic graph convolutional network. A semantic graph reflects the correlations between categories. Based on the semantic graph, graph convolutional networks update the semantic features of all categories. The optimized semantic features are finally aggregated with visual features to guide classification.

\begin{figure}[t]
\begin{center}
   \includegraphics[width=0.6\linewidth]{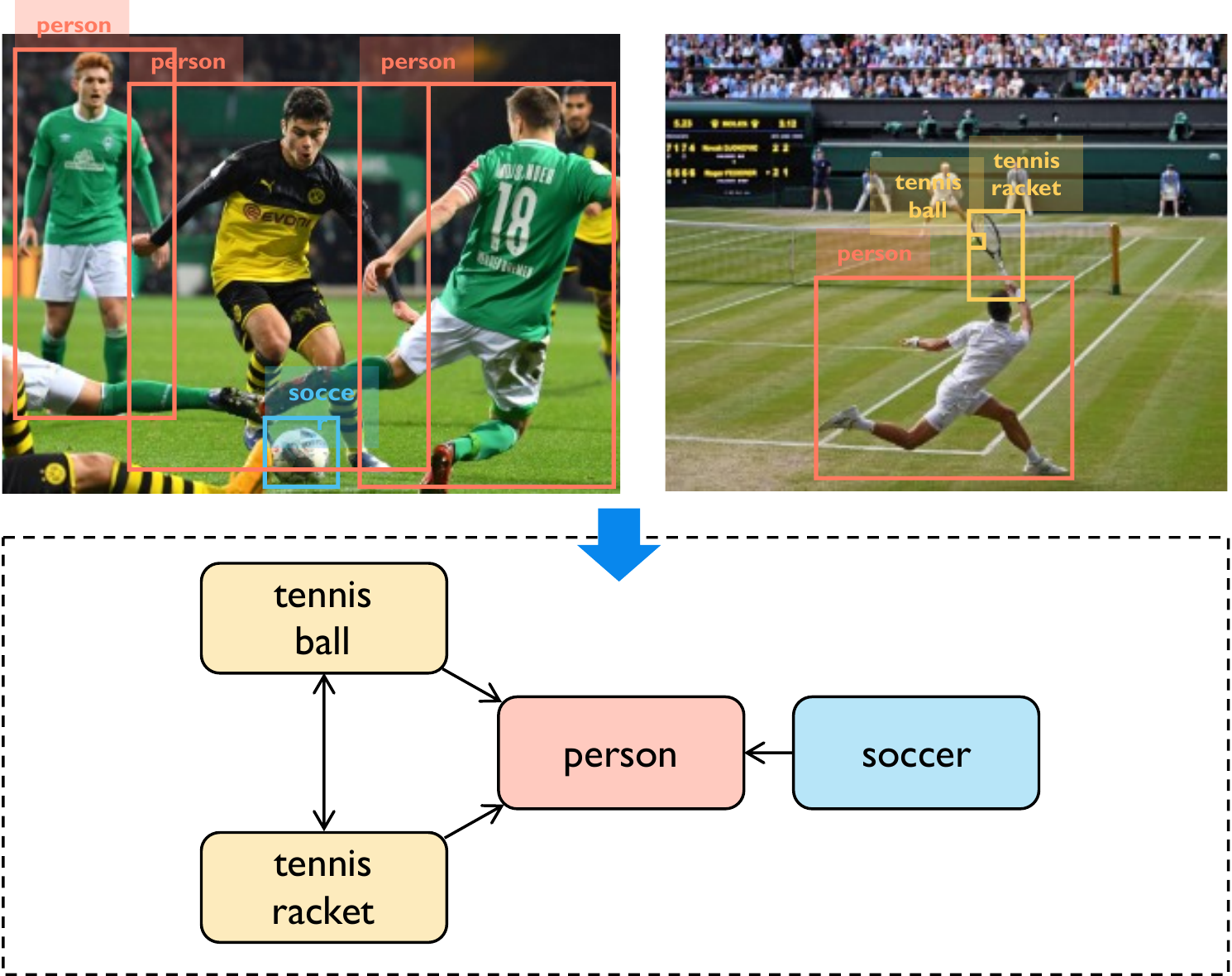}
\end{center}
   \caption{An example of category relationships. The related label graph is a digraph, arrows represent the asymmetric dependencies between categories. ``\emph{soccer} $\to$ \emph{person}" means that when \emph{soccer} appears, \emph{person} appears with high probability.}
\label{fig:labeldependency}
\vspace{-3mm}
\end{figure}

\textbf{Semantic graph construction.} \label{graphconstruction}
We build a semantic graph based on the correlations between all the categories. The corresponding objects of the highly correlated categories often appear jointly, which provides a hint for object detection. Therefore, appropriately leveraging the correlations could lead to better classification and detection results.

For each of the fully-supervised and the weakly-supervised datasets, we build a intra-dataset graph by the co-occurrence of the categories to model the dependences between categories. Here each node models a unique category and each edge reflects the dependence between two categories. Note that the relationship between categories is not bidirectional, such as the~\emph{soccer} and the~\emph{person} in Figure \ref{fig:labeldependency}. Therefore, the dependence is asymmetric and the intra-dataset graph is a digraph. To define the connectivities for each graph, we first compute a graph transition matrix  based on the  co-occurrence of the categories and then apply thresholding to obtain the graph adjacent matrix. For example, the graph transition matrix for the fully-supervised dataset is $\Pj_f \in \mathbb{R}^{C_f \times C_f}$. The $(i,j)$th element of $\Pj$ is the co-occurrence probability between the $i$th and the $j$th categories; that is, $\left(\Pj_f\right)_{ij} = {M_{ij}}/{M_i},$ where $M_{i}$ is the number of images with the $i$th category, and $M_{ij}$ is the number of images with both the $i$th and the $j$th categories. Since the graph transition matrix $\Pj_f$ is mostly a full matrix, which could be noisy and cause expensive computation for the subsequent procedures,  we introduce a threshold $\tau$ to obtain a binary graph adjacent matrix $\Adj_f \in \mathbb{R}^{C_f \times C_f}$. The $(i,j)$th element of the graph adjacent matrix, $\left(\Adj_f\right)_{ij} = 0$ when $\left(\Pj_f\right)_{ij} <  \tau$ and 1, otherwise. Similarly, we can construct the graph adjacent matrix $\Adj_w \in \mathbb{R}^{C_w \times C_w}$ for the weakly-supervised dataset. 

To further capture semantic relations between categories across two datasets, we introduce inter-dataset edges to connect two intra-dataset graphs, forming a bipartite graph $\B \in \R^{C_f \times C_w}$. Each edge weight can be obtained by the cosine similarity as well as the hand-crafted design. For the similarity-based inter-dataset edges The edge weight between the $i$th node in the fully-supervised dataset  and the $j$th node in the weakly-supervised dataset is
	\begin{equation*}
	\B_{ij} = \frac{\exp( {\rm sim}(\vv_i^f, \vv_j^w))}{\sum_{j} \exp( {\rm sim}(\vv_i^f, \vv_j^w))},
	\end{equation*}
where ${\rm sim}(\cdot, \cdot)$ is the cosine similarity, $\vv_i^f$ and $\vv_j^w$ are the semantic features of the $i$th node in the fully-supervised graph and the $j$th node in the weakly-supervised graph respectively.  For the hand-crafted inter-dataset edges, we consider binary edge weights to capture the logical relationships between categories. When two nodes in the respective graphs have the subclass or inclusion relation, the corresponding edge weight is set to be $1$, else $0$. For example, \emph{pedestrain} is a subclass of \emph{person}, so the edge weight between \emph{pedestrain} and \emph{person} is 1.


\begin{figure}[t]
\begin{center}
   \includegraphics[width=\linewidth]{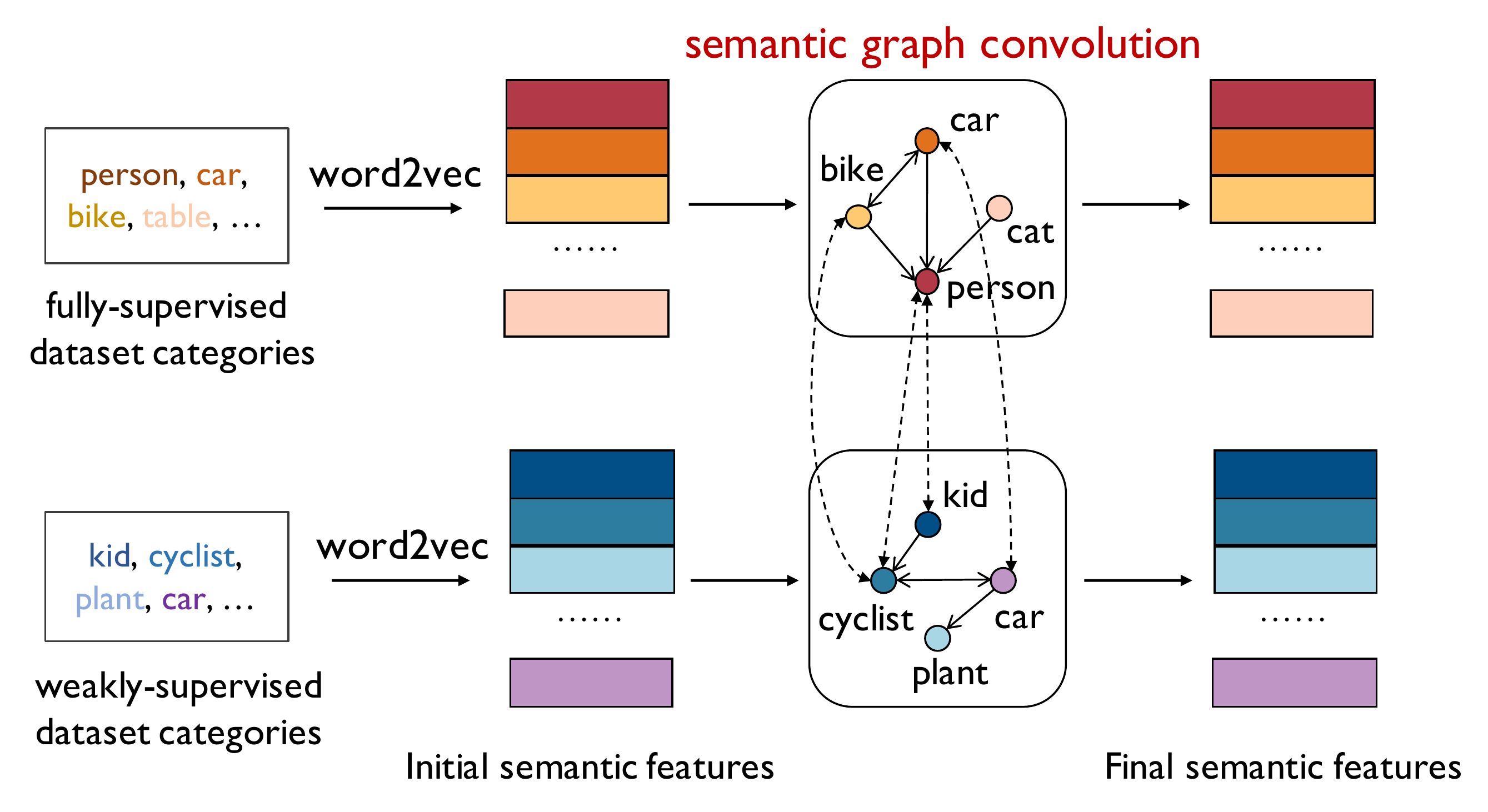}
\end{center}
   \caption{Details of the semantic graph convolutional network. The fully-supervised label graph and the weakly-supervised label graph share the same GNN weights and transmit related semantics between two graphs by inter-dataset graph connections.}
\label{fig:sgm}
\vspace{-3mm}
\end{figure}

\textbf{Network architecture.} \label{na}
Figure \ref{fig:sgm} illustrates the detailed architecture of the semantic graph convolutional network. We first use word2vec pretrained on wiki corpus to initialize the semantic feature for each category,  incorporating semantic information as well as linguistic knowledge. Let $\Hh_f^{(0)} \in \mathbb{R}^{C_f \times k}$ and $\Hh_w^{(0)} \in \mathbb{R}^{C_w \times k}$ be the semantic features of all the categories for the fully-supervised and weakly-supervised datasets, respectively, where $C_f$ is the number of categories of the fully-supervised dataset, $C_w$ the number of categories of the weakly-supervised dataset, and $k$ the dimension of word2vec.  We then use two graph convolution layers~\cite{gcn} to update the semantic features based on the intra-dataset graphs. For the fully-supervised dataset, the $i$th graph convolution layer takes the semantic feature from the previous layer as the input and outputs the latent feature,
$ \Z_f^{(i+1)} = {\rm GCN}(\Adj_f, \Hh_f^{(i)}), $
Similarly, we can obtain the latent feature $\Z_w^{(i+1)}$ for the weakly-supervised dataset. Next, we update the semantic features of the categories in both datasets via the inter-dataset graph:
\begin{equation*}
	\begin{split}
			\Hh_f^{(i+1)}& = \Z_f^{(i+1)} + \sigma(\B \Z_w^{(i+1)} \W_f^{(i)}), \\
			\Hh_w^{(i+1)} &= \Z_w^{(i+1)} + \sigma(\B^T \Z_f^{(i+1)} \W_w^{(i)}),
	\end{split}
\end{equation*}
where $\B \in \mathbb{R}^{C_f \times C_w}$ reflects the inter-dataset graph, $\W_*^{(i)}$ is a trainable weight matrix for each dataset, and $\sigma$ is the activation function. The semantic graph convolutional network outputs the final semantic features for both datasets,  $\Hh_f \in \mathbb{R}^{C_f \times d}$ and $\Hh_w \in \mathbb{R}^{C_w \times d}$, where $d$ is the feature dimension, which is the same as the feature dimension of the visual features $\F_*^o$ in each of two student branches of the double-supervision mean teacher network.

\textbf{Fusion with double-supervision mean teacher network.} Both semantic features are consumed by the classification heads of two student branches. For example, in the fully-supervised student branch, we aggregate the semantic features and the visual features to obtain the final visual feat ures for classification; that is,
$ \check{\F}_f^{o} = g(\F_f^{o}\Hh_f^T) + \F_f^{o} \in \mathbb{R}^{r \times d},$ where $r$ is the number of proposal regions,  $g(\cdot)$ is a $1\times 1$ convolution. The fully-supervised student branch follows the same procedure. Experimental results show that the fused feature $\check{\F}_f^{o}$ is more discriminative than the visual feature $\F_f^{o}$; see Table~\ref{tab:map}.

\vspace{-2mm}
\section{Experiments}

\subsection{Experimental Setup}

\textbf{Datasets.} 
We evaluate our method on Pascal VOC 2007 \cite{voc07} and use COCO 2014 \cite{coco2014} as the supplementary public fully-supervised dataset. Pascal VOC dataset contains 20 categories with 24,640 objects in 9,963 images. COCO dataset contains 80 common object categories with 2.5 million labeled instances in 328k images. The categories in Pascal VOC are totally included by COCO dataset. To better simulate different category overlapping conditions, we process COCO dataset to be COCO-$(80-n)^*$ by deleting the annotations of the randomly picked $n$ overlapping categories, where $(80-n)$ is the number of the remaining categories in COCO dataset. Unlike~\cite{eccv2020}, we only remove the \emph{annotations} but not the \emph{images} of the overlapping categories as objects of the non-overlapping categories may appear in the source fully-supervised dataset. We use COCO-$(80-n)^*$ to denote COCO dataset removing all \emph{images} of the overlapping categories.

We further validate on KITTI dataset~\cite{kitti}. It contains 7 categories related to self-driving excluding \emph{misc} and \emph{dontcare}, with only \emph{car} and \emph{truck} in the label set of COCO dataset. The categories have some logical relationships: \emph{pedestrain} and \emph{person\_sitting} in KITTI are sub-categories of \emph{person} category in COCO; \emph{cyclist} is the combination of \emph{person} and \emph{bicycle}. We use these relationships to build hand-crafted inter-dataset edges between two datasets.

\textbf{Evaluation.} We use mean average precision (mAP) to evaluate the detection performance over categories, and CorLoc~\cite{corloc} to measure the localization accuracy.
 
\textbf{Implementation details.} We use Faster R-CNN\cite{fasterrcnn} as our base model and follow the same settings. We use vgg16\cite{vgg16} pretrained on ImageNet as the feature extraction backbone. For Pascal VOC dataset, we use the similarity inter-dataset edges to build the inter-dataset graph. The dimension of the input semantic features is 300. The semantic graph convolutional network applies two graph convolution layers with the hidden feature dimensions 2048 and 4096.

During training, we do image augmentation by horizontally flipping, randomly cropping, and randomly resizing the image between 0.5 and 2. The initial learning rate is 0.002. We adopt an SGD optimizer with 0.9 momentum and 0.0001 weight decay. The teacher branch weights do not participate in back propagation, and are updated by EMA with EMA decay $\alpha=0.999$. The weights to balance losses are $\lambda_{\rm full} = 0.5$, $\lambda_{\rm cons} = 1.0$. Our method is implemented based on Pytorch Faster R-CNN framework\cite{jjfaster2rcnn}. All the models are trained on 4 NVIDIA 1080 Ti GPU. The batch size is $8$. 


\begin{table}[]
\begin{center}
\resizebox{\linewidth}{!}{
\begin{tabular}{@{}lccc@{}}
\toprule
Method & Source Dataset & mAP(\%) & CorLoc(\%) \\ \midrule
\textbf{pure WSOD:} \\
WSDDN Ensemble~\cite{wsddn} & - & 39.3 & 58.0 \\
OICR-Ens+FR~\cite{oicr} & - & 47.0 & 64.3\\
PCL-Ens+FR~\cite{pcl} & - & 48.8 & 66.6\\
WSOD2~\cite{wsod2} & - & 53.6 & 69.5\\
CASD~\cite{casd} & - & 56.8 & 70.4 \\
\hline
\textbf{WSOD with transfer:}\\
MSD-Ens~\cite{msd} & ILSVRC2013-180 & 51.1 & 66.8\\
OICR+UBBR~\cite{ubbr} & COCO-60$^*$ & 52.0 & 47.6\\
Boosting~\cite{eccv2020} & COCO-60 & 55.2 & 72.4\\
Boosting$^*$~\cite{eccv2020} & COCO-60$^*$ & 57.8 & 73.6\\
\hline
\textbf{Ours:} \\
CaT$_0$ & COCO-60 & 58.0 & 73.8  \\
CaT$_0^*$ & COCO-60$^*$ & 59.2 & 75.9  \\
CaT$_5$ & COCO-65 & \textbf{63.5} & \textbf{80.3}  \\
\hline
\textbf{FSOD:} \\
Faster R-CNN~\cite{fasterrcnn} & - & 69.6 & 94.3\\ \bottomrule
\end{tabular}}
\end{center}
\caption{Comparison of our method on Pascal VOC 2007 test set to FSOD method and the state-of-the-art WSOD methods in terms of mAP (\%) and CorLoc (\%). Our method achieves better performance than the previous WSOD methods. }
\label{tab:main}
\end{table}

\begin{table}[]
\begin{center}
\resizebox{0.7\linewidth}{!}{
\begin{tabular}{@{}c|cc|cccc@{}}
\toprule
Method & Sim & HC & mAP(\%) & CorLoc(\%) \\ \midrule
WSDDN & $\times$ & $\times$ & 64.5 & 26.7  \\
CaT & $\checkmark$ & $\times$ & 83.8 & 56.8 \\
CaT & $\times$ & $\checkmark$ & 80.6 & 51.5 \\
CaT & $\checkmark$ & $\checkmark$ & 79.5 & 58.9 \\
\bottomrule
\end{tabular}}
\end{center}
\caption{KITTI test 2D detection mAP (\%) and CorLoc (\%) with different inter-dataset edges, with `Sim' for the similarity inter-dataset edges, and `HC' for the hand-crafted inter-dataset edges. Our model has good domain adaptation ability and the similarity inter-dataset edges can lead to better performance.}
\label{tab:kitti}
\vspace{-5mm}
\end{table}

\begin{figure*}[t]
\begin{center}
  \subfigure[performance with various  $n$]{\includegraphics[width=0.245\linewidth]{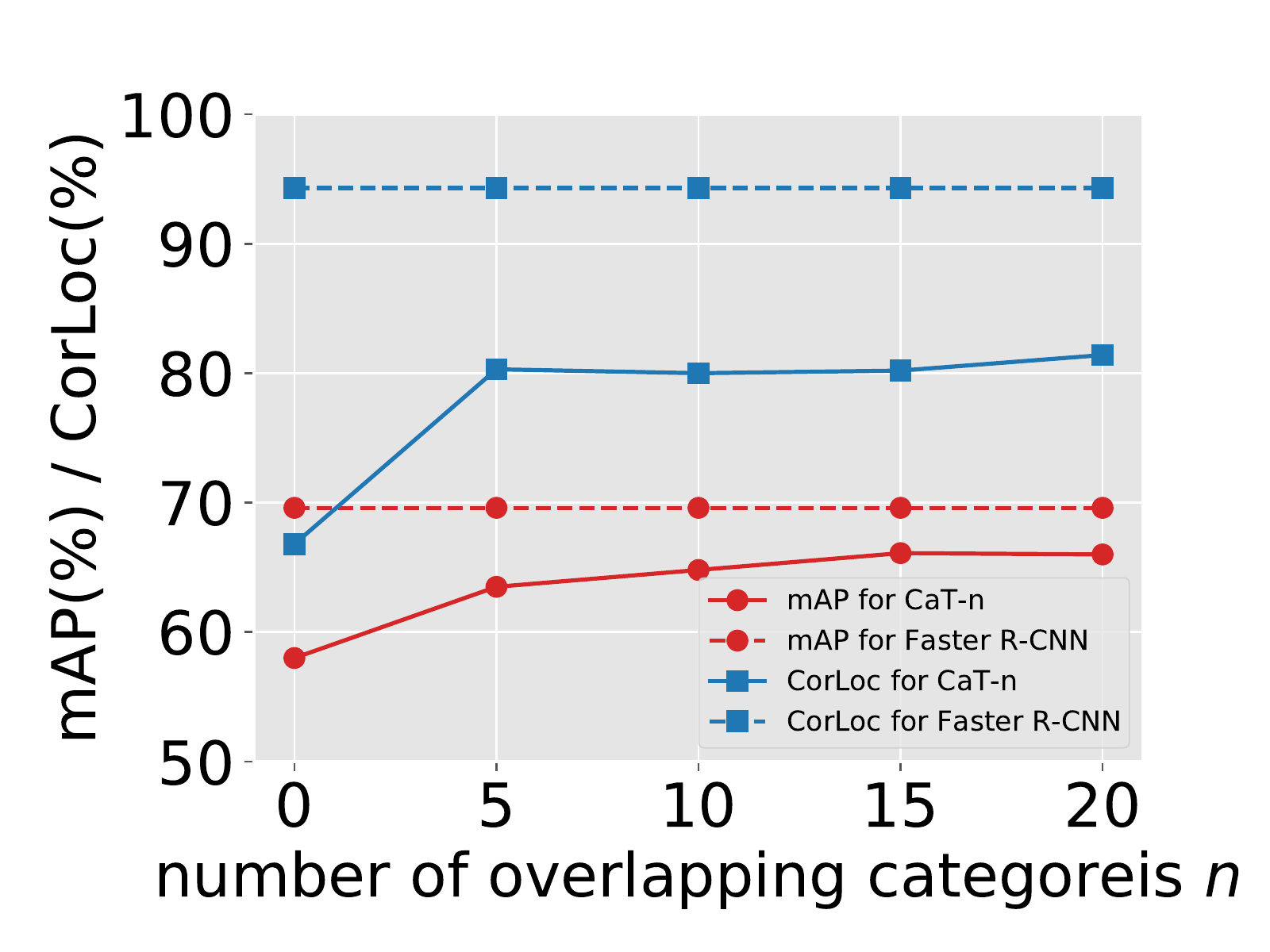}}
  \subfigure[performance with various  $\lambda_{\rm full}$]{\includegraphics[width=0.245\linewidth]{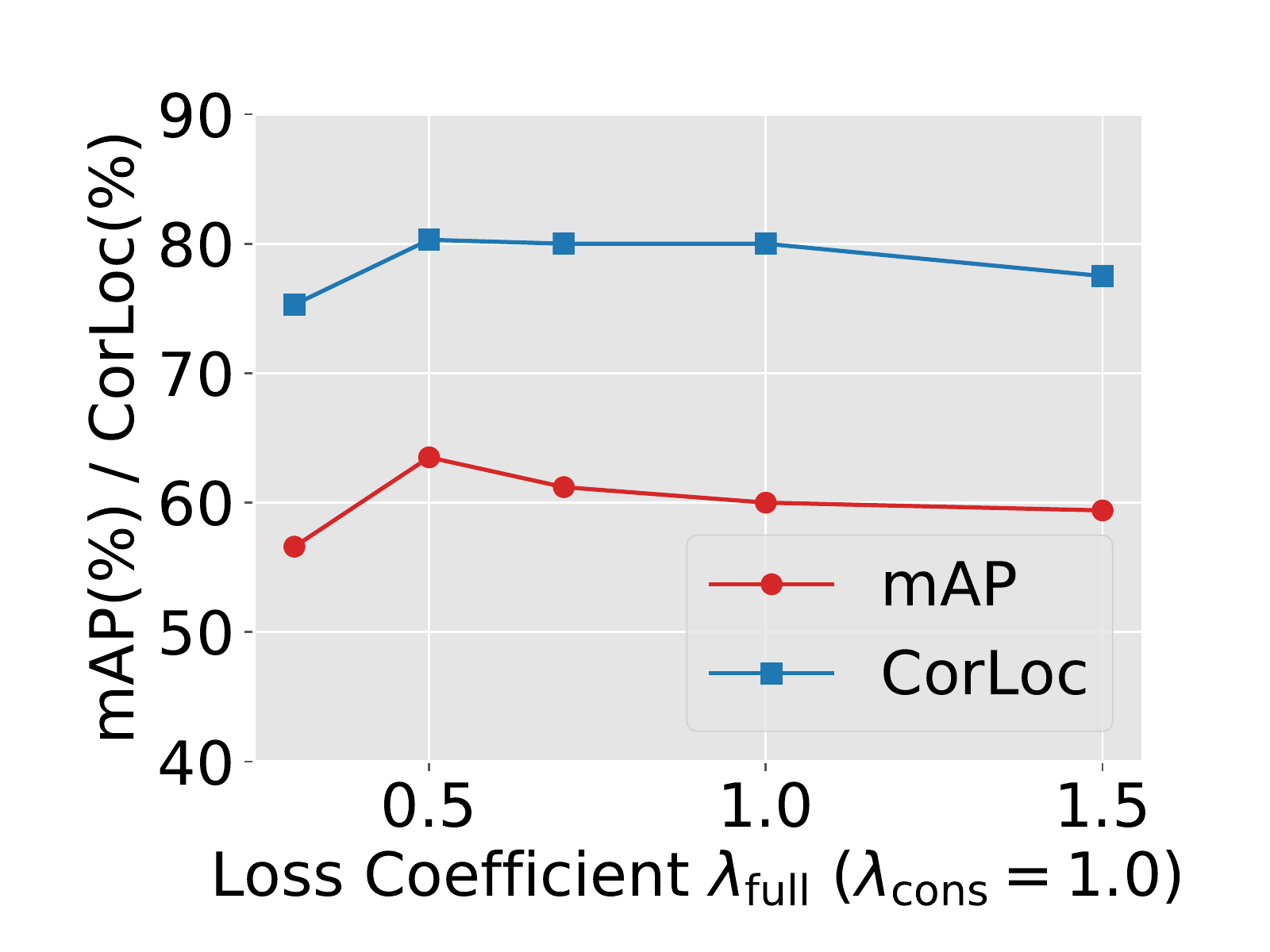}}
  \subfigure[performance with various  $\lambda_{\rm cons}$]{\includegraphics[width=0.245\linewidth]{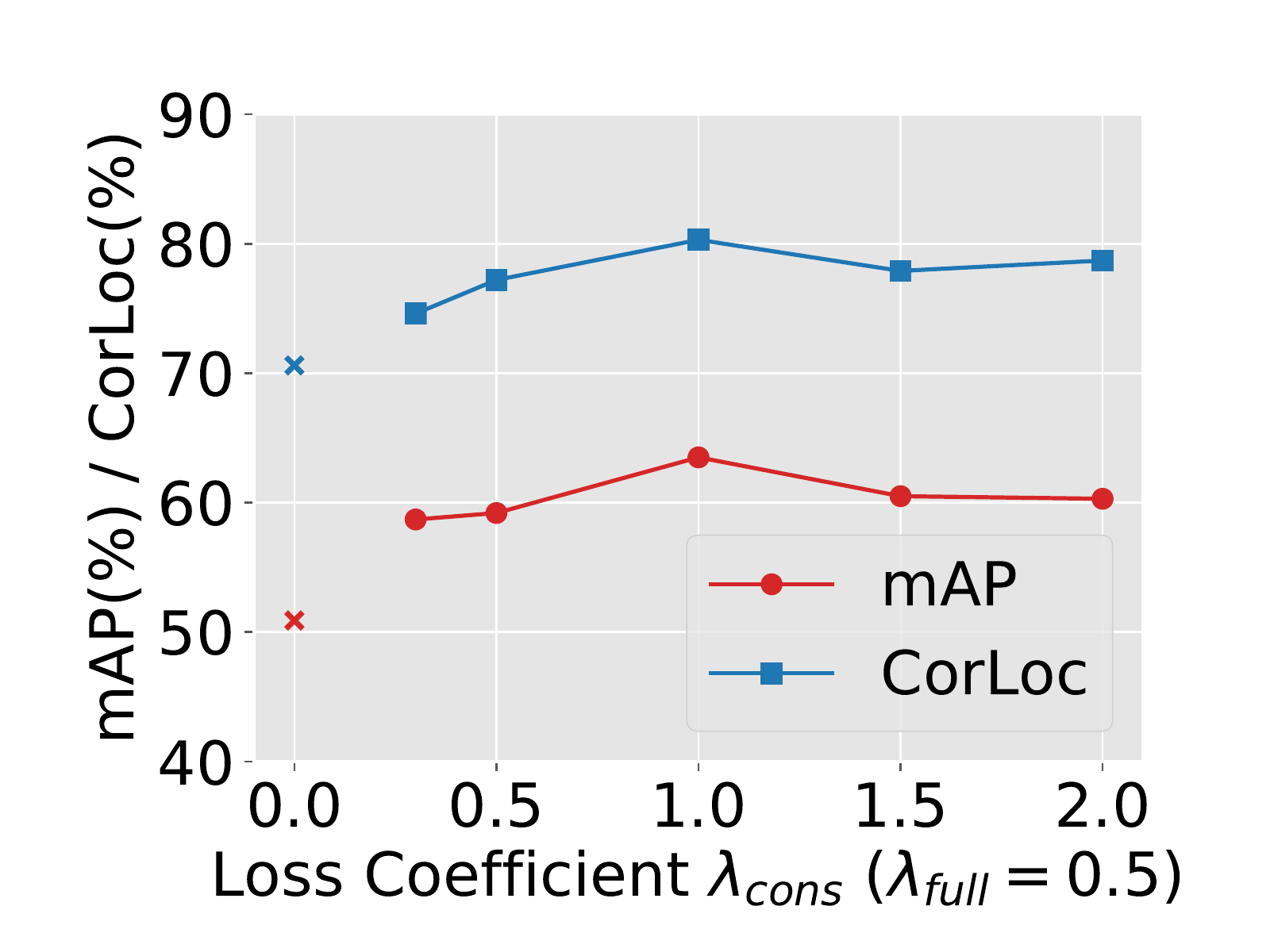}}
  \subfigure[performance with various  $\tau$]{\includegraphics[width=0.245\linewidth]{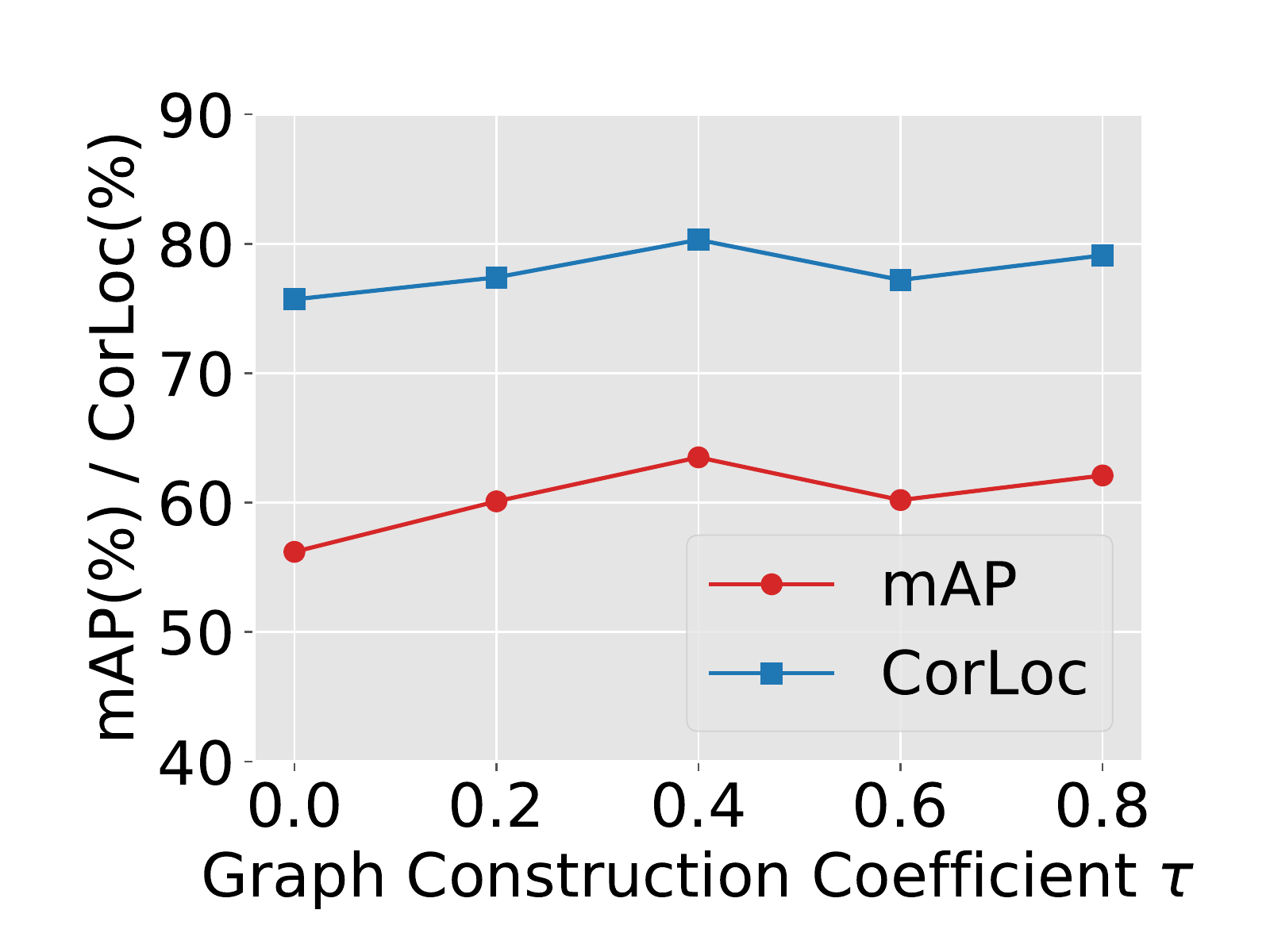}}
\end{center}
   \caption{Ablation study of the number of overlapping category $n$, the balance weight for the fully-supervised loss $\lambda_{\rm full}$, the balance weight for the consistency loss $\lambda_{\rm cons}$, and the graph construction threshold $\tau$. The performance of our method improves with the increase of $n$; and our method is robust to $\lambda_{\rm full}$, $\lambda_{\rm cons}$, and $\tau$.}
\label{fig:ablation}
\end{figure*}

\begin{table*}[]
\begin{center}
\resizebox{\linewidth}{!}{
\begin{tabular}{@{}c|cc|cccccccccccccccccccc|c@{}}
\toprule
& \multirow{2}{*}{DSMT} & \multirow{2}{*}{SGCN} & \multirow{2}{*}{aero} & \multirow{2}{*}{bike} & \multirow{2}{*}{bird} & \multirow{2}{*}{boat} & \multirow{2}{*}{bottle} & \multirow{2}{*}{bus} & \multirow{2}{*}{car} & \multirow{2}{*}{\textbf{cat}} & \multirow{2}{*}{chair} & \multirow{2}{*}{cow} & \multirow{2}{*}{\textbf{table}} & \multirow{2}{*}{\textbf{dog}} & \multirow{2}{*}{horse} & \multirow{2}{*}{\textbf{mbike}} & \multirow{2}{*}{person} & \multirow{2}{*}{\textbf{plant}} & \multirow{2}{*}{sheep} & \multirow{2}{*}{sofa} & \multirow{2}{*}{train} & \multirow{2}{*}{tv} & mAP(\%) \\ 
& &  &  &  &  &  &  &  &  & &  &  &  &  &  &  &  &  &  &  &  &  & CorLoc (\%) \\\hline
\multirow{2}{*}{(A)} & \multirow{2}{*}{$\times$} & \multirow{2}{*}{$\times$} & 29.1 & 49.3 & 31.0 & 25.5 & 25.2 & 40.4 & 63.2 & 34.8 & 26.5 & 39.0 & 0.2 & 29.7 & 41.1 & 53.7 & 33.9 & 27.5 & 36.5 & 33.5 & 32.5 & 51.3 & 35.2 \\
 & &  & 58.5 & 73.2 & 51.6 & 44.3 & 42.9 & 61.8 & 78.3 & 48.8 & 40.0 & 70.1 & 0.0 & 51.5 & 62.4 & 72.5 & 57.3 & 51.6 & 71.4 & 51.3 & 52.1 & 70.98 & 55.5\\ \hline
\multirow{2}{*}{(B)}  &\multirow{2}{*}{$\checkmark$} & \multirow{2}{*}{$\times$} & 72.6 & 49.2 & 54.9 & 40.2 & 49.6 & 79.0 & 80.4 & 69.1 & 43.3 & 75.5 & 30.6 & 65.9 & 71.1 & 59.3 & 67.1 & 31.5 & 68.8 & 59.7 & 70.8 & 66.9 &  60.3\\
 & &  & 87.8 &	69.6 &	78.2 &	64.2 &	71.3 &	90.2 &	92.3 &	82.2 &	60.2 &	92.9 &	58.7 &	81.8 &	84.6 &	77.3 &	79.6 &	58.3 &	91.8 &	73.8 &	80.3 &	82.0 & 77.8\\
\hline
\multirow{2}{*}{(C)}  & \multirow{2}{*}{$\checkmark$} & \multirow{2}{*}{$\checkmark$} & 74.0 & 70.7 & 60.0 & 31.1 & 50.0 & 75.9 & 82.0 & 70.7 & 32.8 & 74.3 & 69.5 & 70.2 & 69.5 & 77.0 & 37.5 & 45.8 & 67.0 & 61.1 & 72.4 & 68.0 & 63.0 \\ 
 & & & 87.3 &	84.4 &	80.3 &	59.1 &	71.3 &	89.1 &	91.7 &	80.7 &	52.5 &	92.9 &	86.2 &	84.5 &	85.0 &	92.3 &	62.6 &	70.1 &	89.8 &	73.5 &	81.9 &	84.7 & 80.0\\
\bottomrule
\end{tabular}}
\end{center}
\caption{The effectiveness of the double-supervision means teacher network (DSMT) and the semantic graph convolutional network (SGCN) in terms of mAP (\%) and CorLoc (\%). The overlapping categories between COCO-65 and Pascal VOC are bold. Both DSMT and SGCN are effective for detection performance and localization accuracy.}
\label{tab:map}
\vspace{-3mm}
\end{table*}

\vspace{-2mm}
\subsection{Comparison with State-of-the-Art Methods}
\textbf{Results on Pascal VOC 2007.} We compare our method with the state-of-the-art approaches on Pascal VOC 2007 dataset, including (1) \emph{pure WSOD methods}: WSDDN~\cite{wsddn}, OICR~\cite{oicr}, PCL~\cite{pcl}, WSOD2~\cite{wsod2}, CASD~\cite{casd}. These approaches are trained without any supplementary data. (2) \emph{WSOD methods with knowledge transfer}: MSD~\cite{msd}, OICR+UBBR~\cite{ubbr}, Boosting\cite{eccv2020}. These methods transfer knowledge from an additional source dataset with bounding box annotations. As for the source dataset, the COCO-$(80-n)$ dataset removes the \emph{annotations} and the COCO-$(80-n)^*$ dataset removes the \emph{images} of the $n$ overlapping categories in the original COCO dataset; ILSVRC2013-180$^*$ is the ILSVRC2013 dataset removing all the images of the 20 categories contained by the Pascal VOC dataset. (3) fully-supervised method: we also compare our method with Faster R-CNN~\cite{fasterrcnn} trained with fully annotations on the weakly-supervised dataset, as in~\cite{eccv2020,pda,pcl}.

Table \ref{tab:main} compares our approach with the previous state-of-the-art approaches on the Pascal VOC test set in terms of mAP and CorLoc. We use CaT$_n$ to denote our method trained with the fully-supervised dataset COCO-$(80-n)$, and use CaT$_n^*$ to denote our method trained with the fully-supervised dataset COCO-$(80-n)^*$.

We first evaluate our method using COCO-60 or COCO-60$^*$ as the source fully-supervised dataset. For pure WSOD approaches, CaT$_0$ outperforms the state-of-the-art method CASD by 1.2\% mAP and 3.4\% CorLoc; and CaT$_0$ improves the WSDDN baseline by 18.7\% mAP and 15.8\% CorLoc. For WSOD approaches with transfer, CaT$_0$ outperforms the state-of-the-art method Boosting~\cite{eccv2020} by 2.8\% mAP and 1.4\% CorLoc with COCO-60 as source; and CaT$_0^*$ outperforms Boosting$^*$~\cite{eccv2020} by 1.4\% mAP and 2.3\% CorLoc with COCO-60$^*$ as source. CaT$_0^*$ using COCO-60$^*$ as source achieves better performance than CaT$_0$ using COCO-60 as source, as the regions with annotation removed in COCO-60 are treated as background while training, which will reduce the recall rate of CaT$_0$.

We then evaluate our method using COCO-65 as the source fully-supervised dataset. With only 5 overlapping categories between COCO-65 dataset and Pascal VOC dataset, CaT$_{\rm 5}$ gets an increase of 6.7\% mAP and 9.9\% CorLoc over the state-of-the-art pure WSOD method CASD, and significantly reduces the mAP gap between weakly supervised object detection method and the fully-supervised method, like Faster R-CNN!\footnote{The qualitative results are shown in the supplementary.}

\textbf{Results on KITTI.}
We also validate the domain adaptation ability of our method. We use a domain general dataset, COCO, as the source fully-supervised dataset and do weakly-supervised object detection on a domain specific dataset, KITTI. For the semantic graph construction, we can build similarity inter-dataset edges as well as hand-crafted inter-dataset edges according to the logical relationships between the categories in KITTI and COCO datasets. Table \ref{tab:kitti} compares the performance of our methods using different inter-dataset edges and a WSDDN baseline. We see that i) our method significantly outperforms our baseline, WSDDN, validating the domain adaptation ability of our method; and ii) our method achieves the best mAP 80.6\% using the similarity inter-dataset edges and the best CorLoc 58.9\% using both the similarity and the hand-crafted inter-dataset edges, which suggests the similarity inter-dataset edges can lead to better category transfer.

\subsection{Ablation Study}

\textbf{Number of overlapping categories.} Figure~\ref{fig:ablation} (a) shows the effects of the number of the overlapping categories $n$ between the fully-supervised and the weakly-supervised datasets. We see that i) the performance improves with the increase of $n$. The reason is that more bounding box and category training data for the specific overlapping categories improve the detector; and 2) our method outperforms Faster R-CNN trained on only COCO-$(80-n)$ dataset with $n$ overlapping categories, validating the domain adaptation ability of our method. For the following ablation studies, we fix the number of overlapping categories as 5. 

\textbf{Network Components.} To validate the effectiveness of the proposed double-supervised mean teacher network (DSMT) and the proposed semantic graph convolutional network (SGCN), we compare three network settings: (A) a pure weakly-supervised object detector using a traditional mean teacher structure without using any proposed network, which consists of the Faster R-CNN backbone, a WSDDN like weakly-supervised student branch, and a teacher branch which has the same architecture with the student branch; (B) the baseline network with double-supervision mean teacher network (DSMT); and (C) the baseline network with double-supervision mean teacher network (DSMT) and the semantic graph convolutional network (SGCN).   Table~\ref{tab:map} compares the detection performances of these three settings. We see that 
i) Setting (B) outperforms (A) by over $25\%$and $20\%$ in terms of the mAP and CorLoc, respectively, indicating the huge benefits brought by DSMT; and ii) Setting (C) further improves (B), validating the effectiveness of the SGCN.

\textbf{Hyperparameters.} Figure \ref{fig:ablation} (b) - (c) shows the effects of hyperparameters in the lost function, $\lambda_{\rm full}$, $\lambda_{\rm cons}$ and $\lambda_{\rm cons}$, respectively. We see that i) too small or too large $\lambda_{\rm full}$ can both causes the performance drop, and our method achieves the best performance with $\lambda_{\rm full} = 0.5$; and ii) A larger $\lambda_{\rm cons}$ means a stronger regularization on $L_{\rm cons}$. The method achieves the best performance with $\lambda_{\rm full} = 1.0$. Plot (d) shows the effect of the intra-dataset graph construction threshold, $\tau$. We see that a larger $\tau$ leads to a sparser intra-datset graph and $\tau = 0.4$ models a proper density of the graph to achieve the best performance. 

\vspace{-2mm}
\section{Conclusion}
\vspace{-1mm}
This paper studies the weakly-supervised object detection with fully-supervised knowledge transfer. We specifically focus on category transfer; that is, leveraging the category information in a fully-supervised dataset to improve the discriminative ability of the detector. For overlapping category transfer, we propose a double-supervision mean teacher network to gather common category information and bridge the domain gap between two datasets. For non-overlapping category transfer, we propose a semantic graph convolutional network to gather semantic features by propagating semantic information between correlated categories. The advantages of the proposed networks lead to a new state-of-the-art on PASCAL VOC 2007.


{\small
\bibliographystyle{ieee_fullname}
\bibliography{egbib}
}

\end{document}